%% file: Template.tex
\documentclass{article}
\usepackage{spconf}
\RequirePackage{amsmath,amssymb,amsthm}
\RequirePackage{mathtools} 
\RequirePackage{mleftright}  
\RequirePackage{bm,bbm} 
\RequirePackage{thmtools}
\RequirePackage{thm-restate}
\usepackage{xcolor,graphicx}
\usepackage[ruled,noend]{algorithm2e}

\graphicspath{{Figures/}}  

\input{papermacros}

\newcommand{\datadist}{\pi}
\newcommand{\dtrain}{\mc{D}_{\mathrm{train}}}
\newcommand{\dtest}{\mc{D}_{\mathrm{test}}}
\newcommand{\Ntrain}{N}
\newcommand{\Ntest}{N'}
\newcommand{\Nmodels}{M}
\newcommand{\mpos}{m^{+}}
\newcommand{\mneg}{m^{-}}
\newcommand{\gapclass}{\mc{M}}

\newcommand{\emprisk}{\hat{R}}
\newcommand{\testacc}{A}
\newcommand{\churn}{C}
\newcommand{\bht}{\underset{\mc{H}_0}{\overset{\mc{H}_1}{\gtrless}}}

\newcommand{\randomness}{\msf{S}}

\newcommand{\trimmed}{\mc{R}}
\newcommand{\distros}{\mc{P}}

\newcommand{\randinit}{\randomness_{\mathrm{init}}}
\newcommand{\randbatch}{\randomness_{\mathrm{batch}}}
\newcommand{\randtrain}{\randomness_{\mathrm{train}}}
\newcommand{\randall}{\randomness_{\mathrm{all}}}

\usepackage[color=red!30,colorinlistoftodos]{todonotes}

\definecolor{sbcolor}{RGB}{0,158,115}

\definecolor{adscolor}{HTML}{6e25d8}

\definecolor{tccolor}{RGB}{230,159,0}

\definecolor{tmcolor}{RGB}{0,114,178}

\title{Robust Nonparametric Hypothesis Testing to Understand Variability in Training Neural Networks}
\name{%
    Sinjini Banerjee$^{\star}$ \qquad %
    Reilly Cannon$^{\dagger}$ \qquad  %
    Tim Marrinan$^{\dagger}$ \qquad  %
    Tony Chiang$^{\dagger}$ \qquad  %
    Anand D.~Sarwate$^{\star}$  \qquad %
    \thanks{The work of S.B. and A.D.S. were supported in part by a Pacific Northwest National Laboratory Program (PNNL) under contract DR00022921. R.C., T.M., and T.C. were partially supported by the Mathematics for Artificial Reasoning in Science (MARS) initiative via the Laboratory Directed Research and Development (LDRD) at PNNL. T.C. and A.D.S. were also partially supported by the Statistical Inference Generates kNowledge for Artificial Learners (SIGNAL) program at PNNL.} 
    }
\address{$^{\star}$Rutgers University \qquad $^{\dagger}$Pacific Northwest National Lab}
\begin{document}

\ninept

\maketitle
\begin{abstract}
Training a deep neural network (DNN) often involves stochastic optimization, which means each run will produce a different model. Several works suggest this variability is negligible when models have the same performance, which in the case of classification is test accuracy.  However, models with similar test accuracy may not be computing the same function. We propose a new measure of closeness between classification models based on the output of the network before thresholding. Our measure is based on a robust hypothesis-testing framework and can be adapted to other quantities derived from trained models.
\end{abstract}
\begin{keywords}
 DNN variability, Nonparametric hypothesis testing, Robust Kolmogorov-Smirnov test
\end{keywords}
\section{Introduction}
\label{sec:intro}


State-of-the-art deep learning models have been remarkably successful in achieving state-of-the-art performance on complex tasks in the fields of healthcare, education, cyber-security, and other important domains. Training these models take significant time, energy, and financial resources. Because training is done with stochastic algorithms for nonconvex optimization, models produced by different training runs in general converge to different solutions. These solutions can be very different, even if they have a similar objective value and test loss. Two models may differ in their predictions on the same test point: this phenomenon is known as churn~\cite{NIPS2016_dc5c768b} and is a cause for model irreproducibility. 

Deep learning models are often continuously retrained as new data arrives. This necessitates algorithmic and architectural changes in state-of-the-art models to improve their performance on new data. The run-to-run variability in training models makes it difficult to conclude if a certain initialization or hyperparameter tuning made a meaningful difference in model performance or if it just ``got lucky'' due to the (unavoidable) presence of randomness in the optimization. Without this knowledge, comparing training configurations to assess if one is better than another becomes a difficult task. 

\begin{figure}[t]
    \begin{center}
    \includegraphics[width=0.4\textwidth]{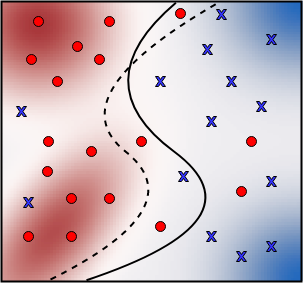}
    \end{center}
\caption{\ninept Illustration of two models that have the same test accuracy but different decision boundaries.}
\label{fig:measures_example}
\end{figure}

Reproductibility in training is an active area of recent work. Gundersen et al.~\cite{gundersen2023sources} identified model initialization, random batch shuffling during mini-batch stochastic gradient descent (SGD), data sampling, and parallel execution over GPUs as some of the major sources of randomness present in the training procedure. Fort et al.~\cite{fort2020deep}, and Bouthillier et al.~\cite{bouthillier2021accounting} showed random data ordering has a smaller effect than random initialization on model performance. Somepalli et al.~\cite{Somepalli} used decision boundaries to characterize the reproducibility of models. Summers and Dinneen~\cite{summers2021nondeterminism}, and Jordan~\cite{jordan2023calibrated} argue that variance matters only during initial conditions of the training procedure.
Most work assesses the impact randomness has on the \emph{test accuracy} of models or \emph{churn} between models, both of which focus on the \emph{decisions} made by predictive (classification) models. Closeness of decisions does not imply closeness of the trained models and high test accuracy does not imply the stability of the learned features or their meaningful contribution to class differentiation~\cite{ho2020avoid}. 

In this paper we try to quantify the variability caused by training in terms of the \emph{network outputs used to make the decision}.  Figure \ref{fig:measures_example} illustrates the difference: the sold and dashed lines represent two decision boundaries between red (circle) and blue (cross). Test accuracy measures incorrect decisions and the churn is given by the region between the two curves. We are interested in the shading, which shows the network output which is thresholded to make decisions.
If we think of the training algorithm as generating a random sample from a function space, we can use other tools to understand model variability. As a first step, instead of looking at whether models make the same number of correct decisions, we can examine the distribution of ``confidence'' values (the \emph{logit gap}) from functions learned by different runs of a fixed deep neural network architecture. 
%
In particular, we use measures derived from nonparametric hypothesis testing~\cite{gibbons1992nonparametric}. We can frame model similarity as asking whether two models generate similar distributions of logit gaps or, with an appropriately defined null, use a goodness-of-fit test to assess whether a particular model is close to the null. Nonparametric goodness-of-fit tests are often sensitive in the large sample regime. To remedy this we use concepts derived from robust statistics~\cite{HuberR:09robustbook}.

\section{Problem Setup}
\label{sec:typestyle}

We focus on a binary classification problem\footnote{While we focus here on binary classification, our framework can be extended to more general prediction problems.} in which data is generated from an (unknown) probability distribution $\datadist$ on the space of feature-label pairs $\cX \times \cY = \R^d \times \{0,1\}$. We interpret a predictive model (e.g.~a neural network) as a function $m \colon \cX \to \R$, where we interpret $m(x)$ as a log posterior probability ratio $m(x) = \log \frac{\P(y = 1 \mid x)}{ \P(y = 0 \mid x) } $.%
 The \emph{prediction} produced by the model is $\hat{y}(x) =  \Indic{ m(x) \ge 0 }$. We can also think of the model as computing a pair of outputs $\mpos(x)$ and $\mneg(x)$ where the posteriors are given by the softmax function: $\P(y = 1 \mid x) = \frac{\exp(\mpos(x))}{ \exp(\mpos(x)) + \exp(\mneg(x))}$. This makes $m(x) =  \mpos(x) - \mneg(x)$,  so we refer to $\mpos(x)$ and $\mneg(x)$ as \emph{logits} and $m(x)$ as the \emph{logit gap}.

A neural network with a given architecture is parameterized by a space of parameters (e.g.~the weights) $\Theta$. Thus for every parameter setting $\theta \in \Theta$ the network computes a function $m(x ; \theta)$ and the neural network defines a family of functions $\gapclass = \{m_{\theta}: \cX \rightarrow \cY \colon \theta \in \Theta \}$. 
Let $[n] = \{1,2,\ldots,n\}$. A training algorithm takes a \emph{training set} $\dtrain = \{ (x_i, y_i) : i \in [\Ntrain] \} \iid \datadist$ and ``learns'' a parameter setting $\theta$ by approximately minimizing an empirical risk $\emprisk(\theta; \dtrain)$ computed over the training data. The output of the training algorithm is a model $m(x; \theta)$ with predictions $\hat{y}(x ;\theta) = \Indic{ m(x;\theta) \ge  0}$.

The optimization algorithms used in NN training are approximate in two ways. First, the risk minimization problem is in general nonconvex so in general they will converge to a local minimum. Second, they are usually stochastic, as in SGD,
which means the estimated parameters $\theta$ are themselves random variables. We can think of an NN training algorithm as sampling from a distribution on $\Theta$ and therefore sampling from the space of functions $\gapclass$.

Since two runs of the training algorithm can produce different functions, it is natural to ask how different these functions are. We can try to answer this using a \emph{test set} $\dtest = \{ (x_j, y_j) \colon j \in [\Ntest] \}$ which we assume is also sampled i.i.d.~from $\datadist$. The \emph{test accuracy} of a model is
    \begin{align}
    \testacc(\theta) = \frac{1}{\Ntest} 
        \sum_{i=1}^{\Ntest} 
            \Indic{ \hat{y}(x_j; \theta) = y_j }.
    \end{align}
Two models $m(\cdot; \theta_1)$ and $m(\cdot; \theta_2)$ which have similar test accuracy make the same number of mistakes. The \emph{churn} is defined by
    \begin{align}
    \churn(\theta_1, \theta_2) = \frac{1}{\Ntest} 
        \sum_{i=1}^{\Ntest} 
            \Indic{ \hat{y}(x_j; \theta_1) \ne \hat{y}(x_j; \theta_2) },
    \end{align}
which is the fraction of training points where the models disagree. Two models with low churn make almost the same decisions (regardless of whether they are correct or not).



Both test accuracy and churn focus on the \emph{predictions} made by models and do not use information about the logit gap function $m(x)$ beyond its sign. Looking at $m(x)$ directly gives us other approaches to assess whether models are similar or not: two models may have similar accuracy and low churn but can have very different logit gaps. Given the test set, our central object of inquiry is the empirical distribution of logit gaps, which we can write as a cumulative distribution function (CDF):
    \begin{align}
    F(x; \theta) = \frac{1}{\Ntest} \sum_{j=1}^{\Ntest} 
        \Indic{ m(x_j; \theta) \le x }.
    \end{align}
With some abuse of notation, we consider $\distros$ to be the set of probability measures on $\R$ and a $F$ as an element of $\distros$.

We consider the following statistical setup. The training algorithm takes the training set $\dtrain$ and uses randomization in several different ways to produce a model. Each run of the training algorithm uses independent randomness: training is therefore sampling parameters $\theta_1, \theta_2, \ldots, \theta_{\Nmodels}$ i.i.d.~from an (unknown) distribution on $\Theta$ induced by the training algorithm. These parameters correspond to $\Nmodels$ i.i.d.~samples $\{ m_k(x) = m(x ; \theta_k) : k \in [\Nmodels]\}$ taking values in $\gapclass$. When applied to the test set, these produce $\Nmodels$ CDFs $\{ F_k(x) = m(x ; \theta_k) : k \in [\Nmodels]\}$.

We can control which sources of randomness are used in the training by altering the training procedure. For example, we can use deterministic initialization or fixed batch ordering. Under these scenarios, we can generate $\Nmodels$ models and ask if the models are more different from each other when we turn on or off the sources of randomness. To measure how different the models are, we use a framework from nonparametric hypothesis testing.

\section{Goodness-of-fit testing for logit gaps}

We can define an expected logit gap function $\bar{m}(x)$ by integrating over the distribution on $\gapclass$ induced by the training algorithm. Given a test set $\dtest$, this produces an expected CDF which we will call $F_0$. If we knew what $F_0$ was, we could assess whether or not a given model $m(x)$ is a representative sample from the training algorithm by testing if the model's CDF $F(x)$ is close to the expected $F_0(x)$. This corresponds to the following one-sided  hypothesis test:
\begin{align}
    \mc{H}_0 &\colon \{ m(x_j) \colon j \in [\Ntest] \} \sim F_0 \\
    \mc{H}_1 &\colon \{ m(x_j) \colon j \in [\Ntest] \} \not\sim F_0
\end{align}
This is a classical nonparametric goodness-of-fit testing problem which can be solved using the Kolmogorov-Smirnov (KS) test~\cite{an1933sulla,smirnov1948table,massey1951kolmogorov}: $d_{k}(F_{0}, F) = \norm{ F_{0}(x) - F(x) }_{\infty}
     \bht \tau$,
where $F(x)$ is the empirical CDF from $m(x)$ and we can set the threshold $\tau$ to achieve the desired error tradeoff.

In large sample settings, the KS test often rejects the null because even small changes in the sample can result in a significant shift in the $L_{\infty}$ norm. Note that we do not expect our empirical samples to look exactly like they were drawn from $F_0$. Ideally, we want a test which allows for some outliers. Furthermore, we do not know what the null hypothesis $F_0$ is, so even computing the test statistic is not possible. To address these issues, we use ideas from \emph{robust statistics}~\cite{HuberR:09robustbook}. 
Given a distribution $P \in \distros$ and $\alpha \in [0,1]$, we can define the set of $\alpha$-contaminated distributions as
    $\trimmed_{\alpha}(P) = \{ (1 - \alpha) P + \alpha Q \colon Q \in \distros \}.$
Alvarez-Esteban et~al.~\cite{alvarez2008trimmed} interpret $\trimmed_{\alpha}(P)$ as the set of so-called $\alpha$-trimmings of $P$:
    \begin{align} \label{eq:def:trimming}
    \trimmed_{\alpha}(P) =  \braces*{ Q \in \distros \colon Q \ll P,\frac{dP}{dQ} \leq \frac{1}{(1 - \alpha)} P-\text{ a.s}. }.
    \end{align}
The advantage of this definition, as shown by del Barrio et al.~\cite{del2020approximate}, is that we can compute, for a given $F$, the closest $L_{\infty}$ approximation to $F_0$ in the set of $\alpha$-trimmings of $F$.
 \begin{align}
    d_{k}(F_{0}, \trimmed_{\alpha}(F)) 
    = \min_{\tilde{F} \in \trimmed_{\alpha}(F)}
        \norm{F_0 - \tilde{F}}_{\infty}.
    \label{eq:trimmings}
\end{align}

The core idea behind trimming is to ask if trimming a small fraction of samples (from the test set) would allow a KS test to accept the null hypothesis. The set $\trimmed_{\alpha}(F)$ of $\alpha$ trimmings of P can be characterized in terms of a ``trimming function'' $h$ and we can efficiently find the optimizing $\tilde{F} \in \trimmed_{\alpha}(F)$ in \eqref{eq:trimmings} by optimizing over these trimming functions. This technique led to a recent work of del Barrio et al.~\cite{del2020approximate} that proposes a \emph{robust KS test} based on trimming.

\section{Using trimming to estimate variability}

Our main contribution is a new way to measure the dissimilarity of trained models. To do this we first develop a two-sample version of the robust KS test discussed in the previous section.

\noindent \textbf{Using a deep ensemble as a base model.}
While we do not know the expected CDF $F_0$, note that for a given test point $(x_j, y_j)$, we can use the empirical mean $\frac{1}{\Nmodels} \sum_{k=1}^{\Nmodels} m_k(x)$ as an estimate of expected logit gap function $\bar{m}(x) = \expect{ m(x) }$ at test point $x$. To evaluate whether a particular model with CDF $F_{\ell}$ is close to $F_0$ we can instead measure its closeness to the \emph{leave-one-out ensemble} (LOOE):
    \begin{align}
    m_{-\ell}(x) &= \frac{1}{\Nmodels-1} 
        \sum_{k \ne \ell} m_k(x)
    \\
    F_{-\ell}(x) &= \frac{1}{\Ntest} 
        \sum_{j = 1}^{\Ntest} \Indic{ m_{-\ell}(x_j) \le x }.
    \label{eq:looe}
    \end{align}  
The LOOE corresponds to a deep ensemble predictor~\cite{lakshminarayanan2017simple,lee2015m,pearce2018highquality}, which has been used to reduce variability in DNN models. 
Taking the average of model "confidences" across independent training runs make them closer to their expected values, lowering variability. Figure \ref{fig: alpha accuracy} shows how the logit gap samples obtained from averaging over all candidate models compare with candidate models in the pool. The ensemble model produces a lower number of samples with small logit gaps (samples with higher uncertainty), and large logit gaps (overconfident samples). 

\noindent \textbf{Proposed algorithm.}
In our robust two-sample KS test, we are comparing two empirical CDFs, a candidate $F_{\ell}$ and the LOOE  $F_{-\ell}$, and asking if they were generated by a common underlying distribution:
    \begin{align}
    \mc{H}_0 &\colon F_{\ell}, F_{-\ell} \text{\ are from the same distribution}
    \\
    \mc{H}_1 &\colon F_{\ell}, F_{-\ell} \text{\ are not from the same distribution}
    \end{align}
We use bootstrap sampling to resample $\Ntest$ independent test points to compute $F_{\ell}$ and $F_{-\ell}$. For a fixed $\alpha$, we calculate the $\alpha$-trimming of $F_{\ell}$ that minimizes the $L_{\infty}$ distance to $F_{-\ell}$:
    \begin{align}
    d_{k}(F_{-\ell}, \trimmed_{\alpha}(F_{\ell})) 
    = \min_{F \in \trimmed_{\alpha}(F_{\ell})}
        \norm{F_{-\ell} - F}_{\infty}.
    \label{eq:twosamp:dist}
    \end{align}
Our hypothesis test is then
    \begin{align}
    d_{k}(F_{-\ell}, \trimmed_{\alpha}(F_{\ell})) \bht \tau.
    \label{eq:robust2sample_ht}
    \end{align}
To set the threshold $\tau$, we observe that if $\tilde{F}$ is the minimizer in \eqref{eq:twosamp:dist}, the two-sample version of the Dvoretzky-Kiefer-Wolfowitz inequality due to Wei and Dudley~\cite{WeiD:12dkw} implies that
  \begin{align}
  \prob*{ 
    \sup_{x} \abs*{ \tilde{F}(x) - F_{-\ell}(x) } > \tau
    }
    \leq C e^{-2 \Ntest \tau^{2}}.
    \end{align}
For $\Ntest > 458$ the leading constant $C$ can be replaced with $2$. In that case, to set the probability of falsely rejecting $\mc{H}_0$ to  $\delta$, we can set the right hand side equal to $\delta$ and solve for 
    $\tau = \sqrt{ \Ntest \ln \frac{2}{\delta} }.$
We use this test in Algorithm \ref{alg:alpha_search} to compute the largest $\alpha$ for which the test fails to reject $\mc{H}_0$. The output $\hat{\alpha}$ is our proposed measure of discrepancy between model $m_{\ell}$ and the LOOE model.

\begin{algorithm}\label{alg:alpha_search}
\caption{ \ninept  Estimate $\hat{\alpha}$ measure}
\KwData{$\dtest$ with $\Ntest > 458$ samples, trained models $\{ m_k \colon k \in [\Nmodels] \}$, model index $\ell \in [\Nmodels]$, threshold $\tau$, resampling number $B$, trimming levels $(\alpha_1, \ldots, \alpha_T)$}
\KwResult{trimming level estimate $\hat{\alpha}$}
\For{$b = 1$ to $B$}{
    Compute $F_{\ell}$ using $\Ntest$ resampled from $\dtest$\;
    Compute $F_{-\ell}$ in \eqref{eq:looe} using $\Ntest$ resampled from $\dtest$\;
    $\msf{Reject} \gets 1$, $\hat{\alpha}_b \gets 0$, $t \gets 1$\;
    \While{$\msf{Reject} = 1$ and $t \le T$}{
        $\alpha \gets \alpha_{t}$\;
        $\mc{H} \gets$ output of \eqref{eq:robust2sample_ht} \;
        \eIf{$\mc{H} = \mc{H}_0$}{
            $\msf{Reject} \gets 0$, $\hat{\alpha}_b \gets \alpha$  
        }{
            $t = t + 1$\;
        }
  }
}
$\hat{\alpha} \gets \frac{1}{B} \sum_{b=1}^{B} \hat{\alpha}_b$\;

   
 \end{algorithm}

\section{Experiments}
\label{sec:experiments}

To illustrate our approach, we used a small convolutional neural network on two classes from the CIFAR-10 dataset~\cite{krizhevsky2009learning}: $\Ntrain = 12000 $ and $\Ntest = 4000$. Our network had two convolutional layers (having $32$ and $16$ features, respectively, with a $3 \times 3$ kernel size) followed by one hidden layer of 64 units and a final layer of 2 units that output the raw logits of the network. The small size allows us to train many more models to explore model training in different scenarios: $\randinit$ with only random initialization, $\randbatch$ with only random batch selection in SGD, $\randtrain$ with only randomly resampled training data, and $\randall$ with all sources of randomness.
For each scenario we trained $\Nmodels = 100$ models for $50$ epochs.

\begin{figure}[htb]
 \begin{minipage}[b]{0.49\linewidth}
  \centering
  \centerline{\includegraphics[width=4.22cm]{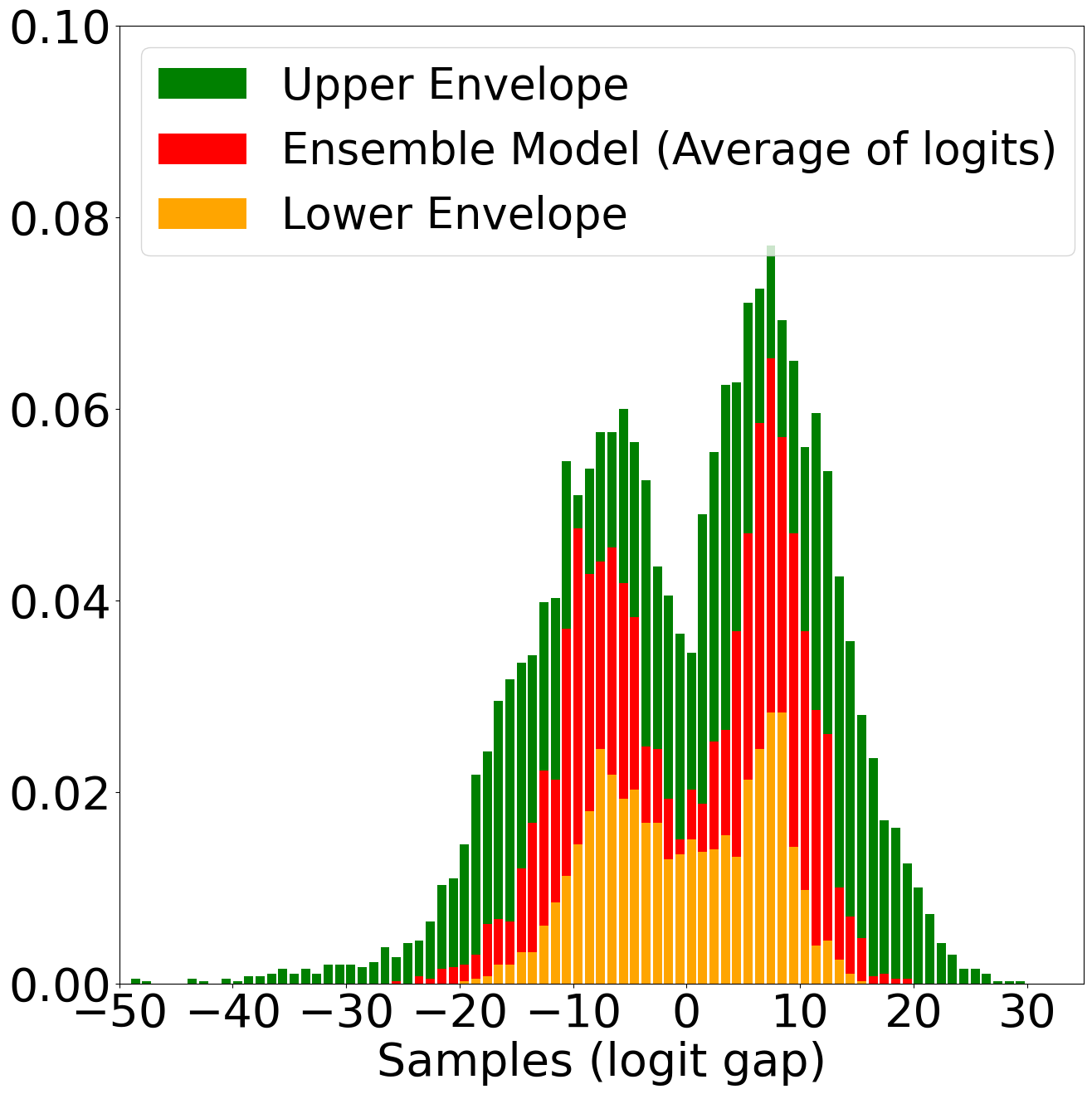}}
\end{minipage}
\begin{minipage}[b]{0.49\linewidth}
  \centering
  \centerline{\includegraphics[width=1.0\textwidth]{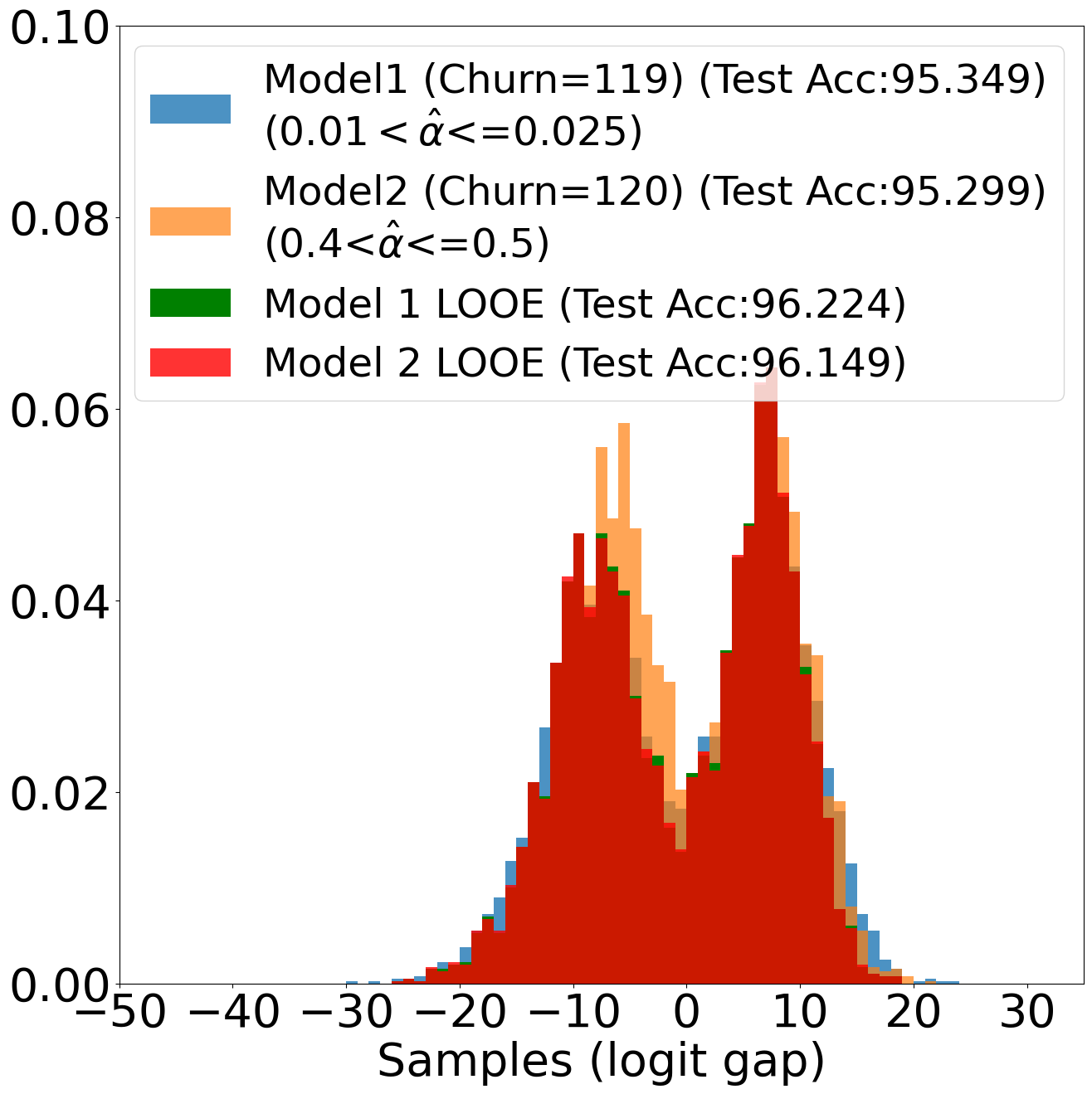}}
\end{minipage}
\caption{ \ninept (Left) Histogram of logit gap from the ensemble model with the upper and lower envelope representing the maximum and minimum probability attained in each bin among individual candidate models. (Right) Histogram plots of logit gap samples from two candidate models and their LOOE models at a fixed epoch.}
\label{fig: alpha accuracy}
\end{figure}

\noindent \textbf{Similar accuracy/churn do not imply small $\hat{\alpha}$.} To show that our measure $\hat{\alpha}$ provides a different understanding of model closeness, Figure \ref{fig: alpha accuracy} shows the logit gap distribution for two models with respect to their LOOE. The churn is computed w.r.t to the LOOE. In general, if models need a small trimming level to be accepted under the null hypothesis they are likely also to have a low churn and high test accuracy.  However, the opposite is not necessarily always true. As seen in Figure \ref{fig: alpha accuracy}, Model 2 has produced more uncertain samples with a higher probability when compared to another candidate model with a similar test accuracy and churn. This could make Model 2 more prone to adversarial attacks. Based on this we observe that a high test accuracy or a low churn is not enough to conclude that one model is a better model than all other candidate models belonging to the same pool. Thus, the recommendation is to choose models that not only have achieved good test accuracy and low churn but are also better representatives of the LOOE. Relying on $\hat{\alpha}$ avoids models that somehow "got lucky" and lets us choose better models that have performed meaningfully well.



\begin{figure}
\begin{minipage}[b]{0.49 \linewidth}
  \centering
  \centerline{\includegraphics[width=4.3cm]{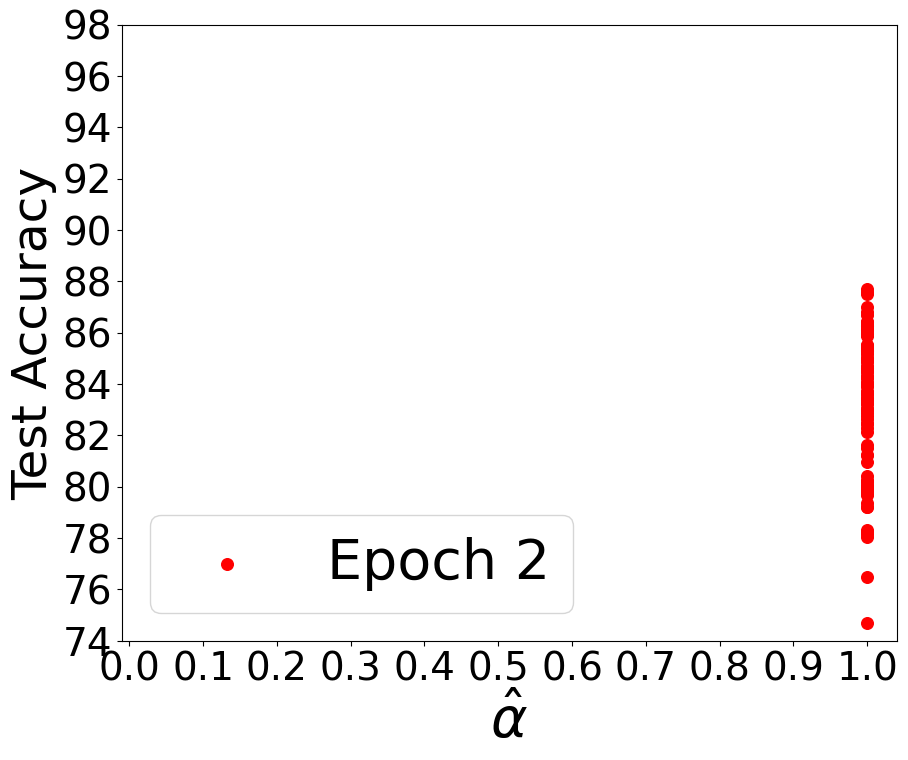}}
\end{minipage}
\begin{minipage}[b]{0.49 \linewidth}
  \centering
  \centerline{\includegraphics[width=4.3cm]{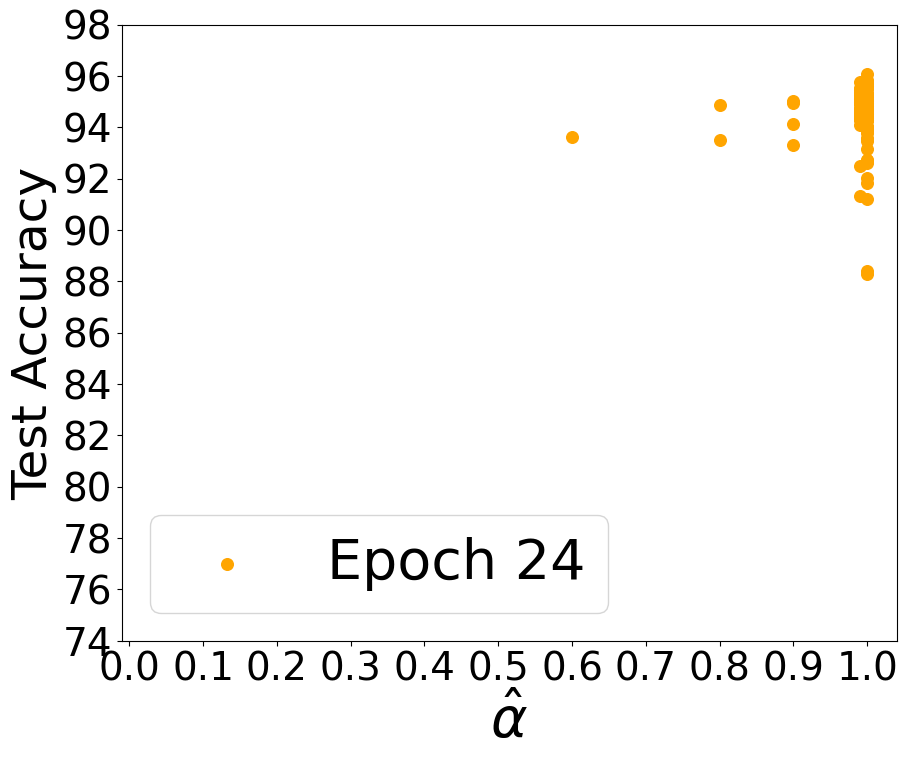}}
\end{minipage}
\begin{minipage}[b]{0.49 \linewidth}
  \centering
  \centerline{\includegraphics[width=4.3cm]{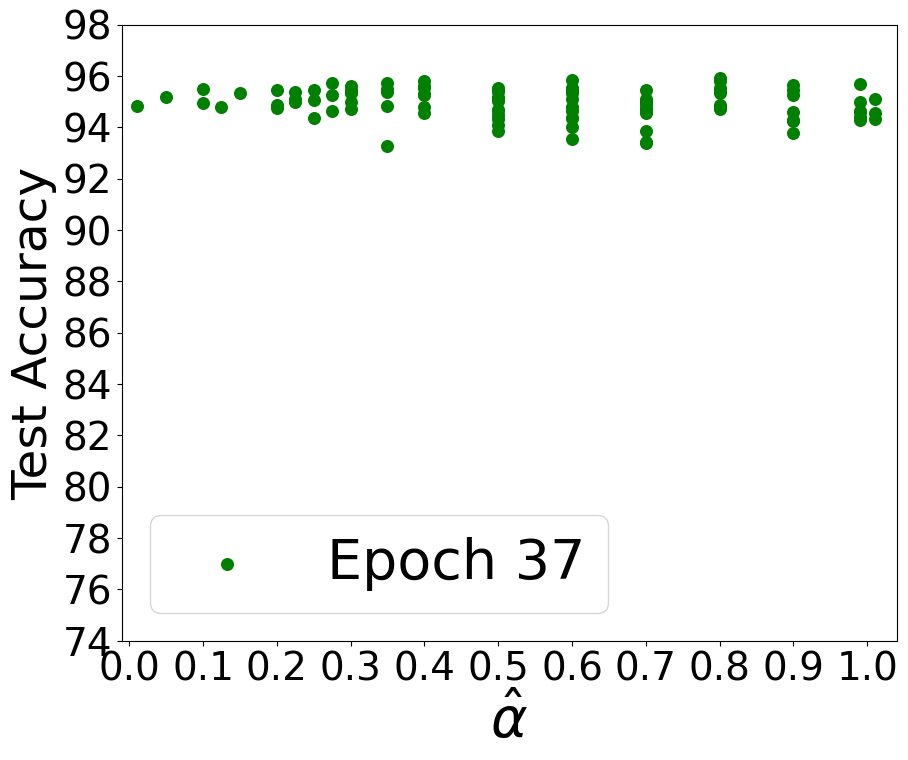}}
\end{minipage}
\begin{minipage}[b]{0.49 \linewidth}
  \centering
  \centerline{\includegraphics[width=4.3cm]{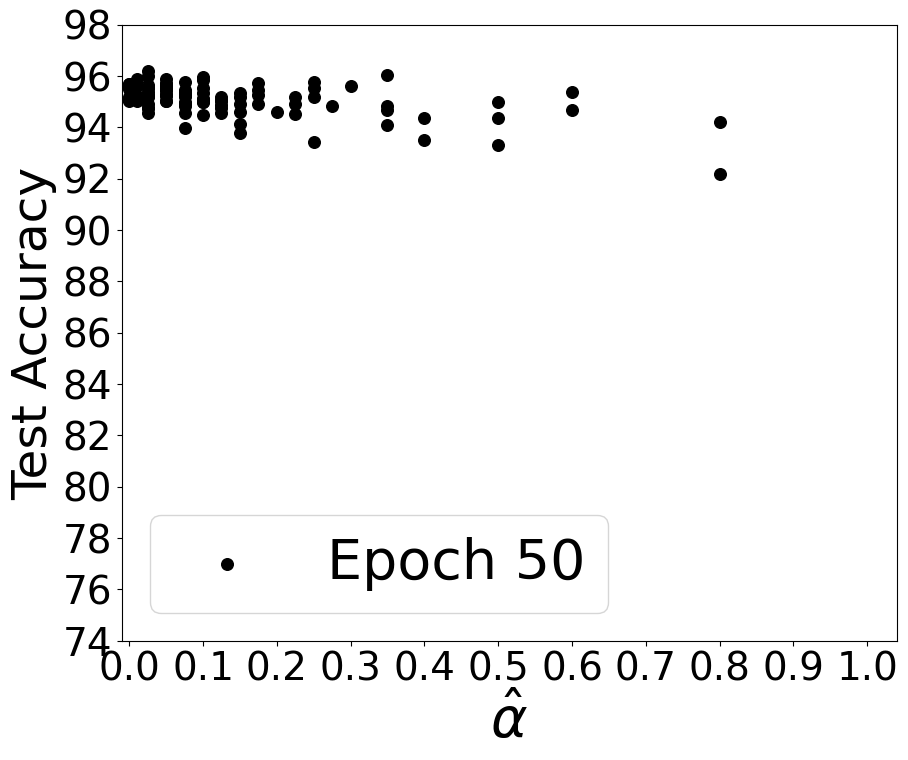}}
\end{minipage}

\caption{ \ninept Scatterplot of $\hat{\alpha}$ vs.~test accuracy   for different epochs under $\randall$. Each model is compared to the LOOE model at epoch 50.}
\label{fig:scatterplot:alpha_v_accuracy}
\end{figure}

\noindent \textbf{Comparing the evolution of $\hat{\alpha}$ and test accuracy.} To understand the difference between $\hat{\alpha}$, test accuracy, and churn, we considered the baseline scenario $\randall$. We trained $\Nmodels = 100$ models for $50$ epochs to examine the evolution of each model over those 50 epochs compared to the LOOE at 50 epochs. We have only included test accuracy plots due to space restrictions but the churn of these models w.r.t to their LOOE also follows a similar behavior. Figure \ref{fig:scatterplot:alpha_v_accuracy} shows four snapshots of the relationship between the test accuracy of a model and $\hat{\alpha}$.

At smaller epochs, although test accuracy improves, $\hat{\alpha}$ remains large for most models. After training the models long enough, we stop seeing a large improvement in test accuracy but candidate models move closer in distribution toward their LOOE model at a higher epoch. This suggests that candidate models evolve to become better representatives of the training algorithm long after reducing test accuracy variability, and $\hat{\alpha}$ is indicative of this evolution.

\noindent \textbf{Using $\hat{\alpha}$ to examine the impact of randomization.} 
Our focus is on two epochs, one very early on in the training (epoch 2) and another after training for a long period of time (epoch 50). At epoch 2, models differing according to $\randinit$ had a higher percentage of models that do not reject the null hypothesis, with a low $\hat{\alpha}$, than $\randtrain$, or $\randbatch$. This would suggest that $\randinit$, contributes the least among individual sources of variance to $\randall$, by a significantly large margin. However, both $\randinit$, and $\randbatch$ seem to contribute less than $\randtrain$ in test accuracy variability at epoch 2. This sensitivity of $\hat{\alpha}$ to model similarity/dissimilarity is lost at higher epochs. At epoch 50, we see a reduction in variability both in test accuracy and $\hat{\alpha}$. This points us to the work done by \cite{jordan2023calibrated}, that individual sources stop contributing independently to the total
variance observed from changing all sources of randomness
if you train the models long enough.

\begin{figure}[!htb]
\begin{minipage}[b]{0.49 \linewidth}
  \centering
  \centerline{\includegraphics[width=4.3cm]{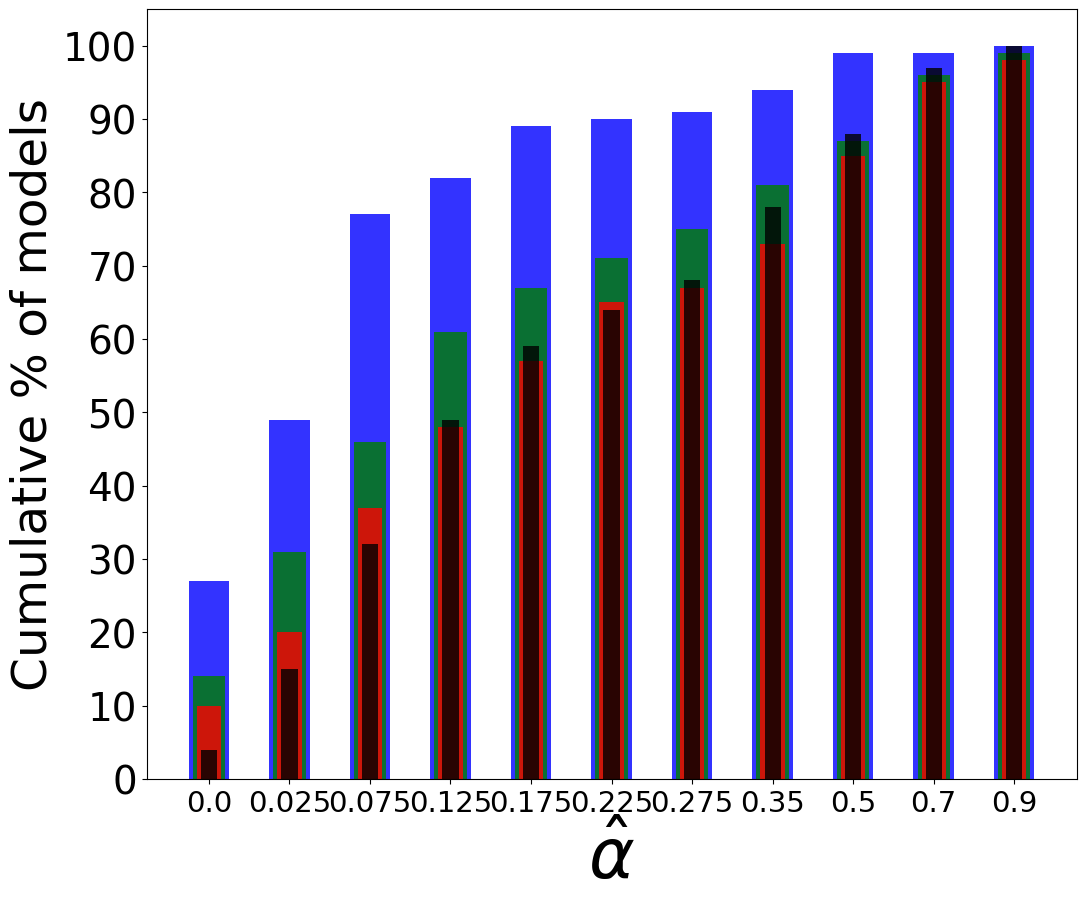}}
\end{minipage}
\begin{minipage}[b]{0.49 \linewidth}
  \centering
  \centerline{\includegraphics[width=4.1cm]{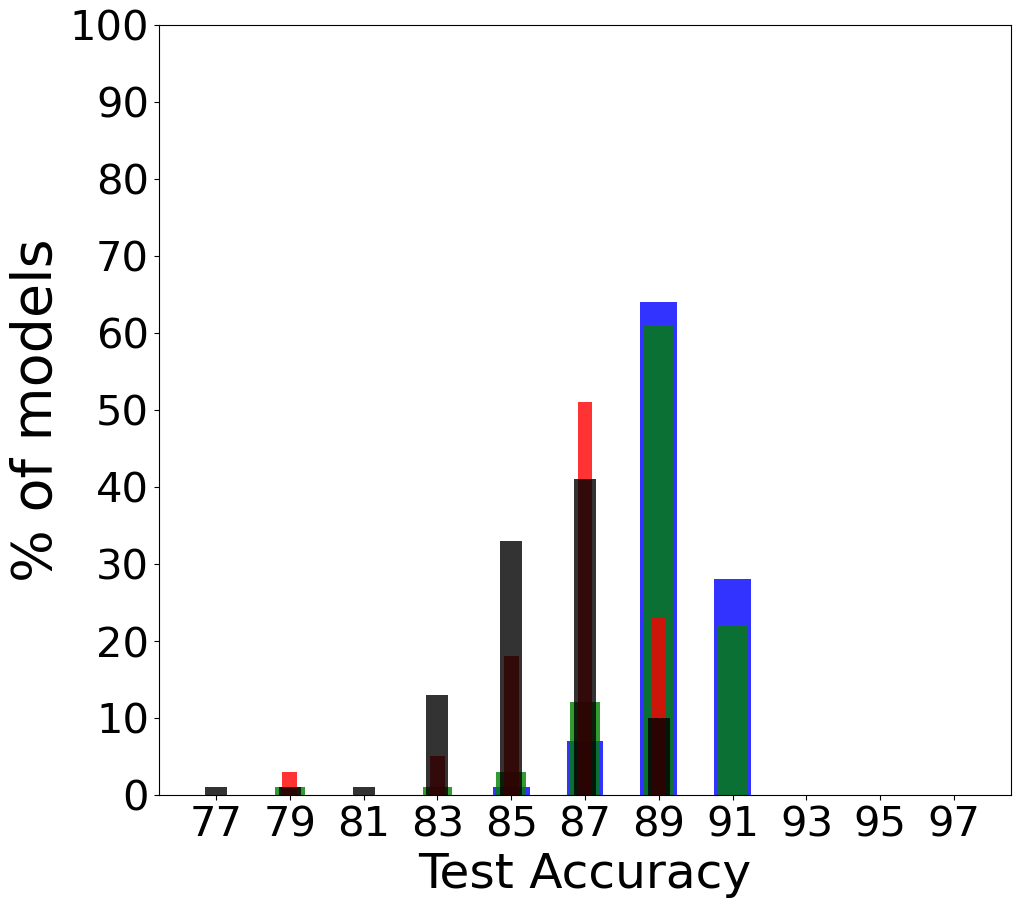}}
\end{minipage}
\begin{minipage}[b]{0.49 \linewidth}
  \centering
  \centerline{\includegraphics[width=4.3cm]{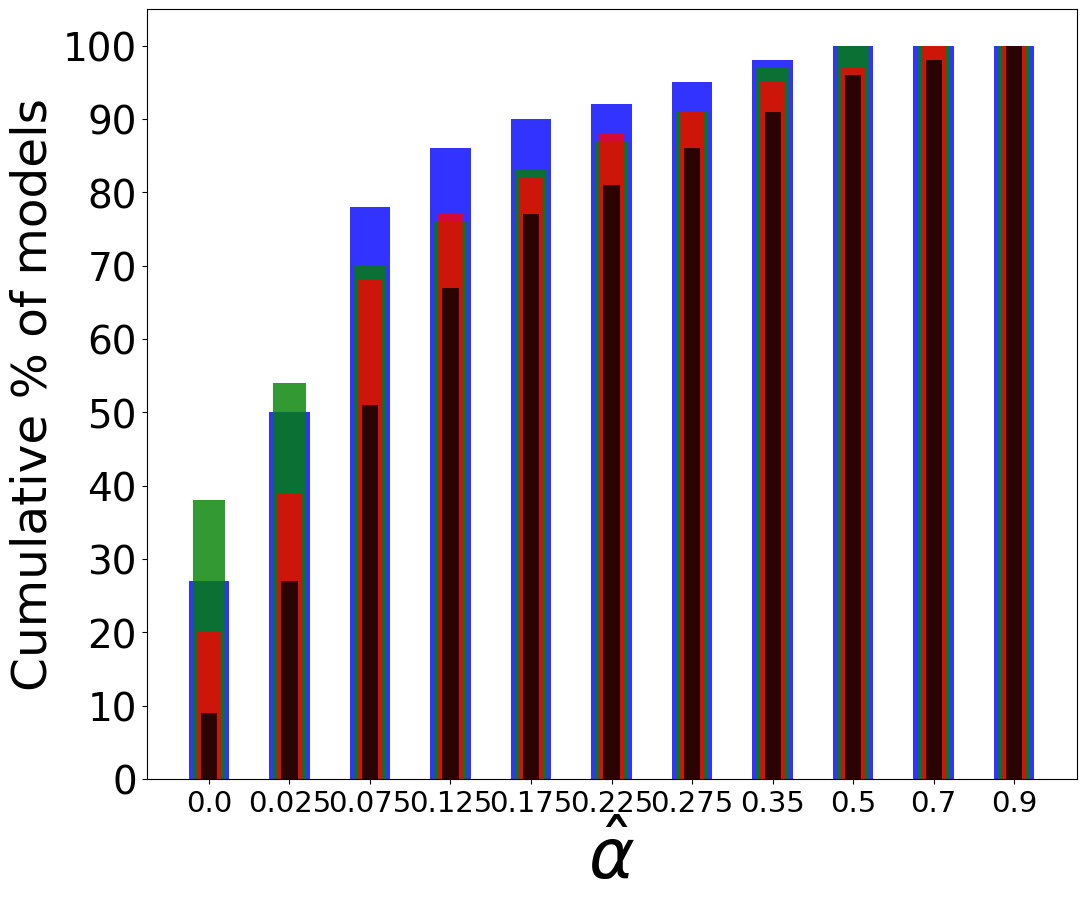}}
\end{minipage}
\begin{minipage}[b]{0.49 \linewidth}
  \centering
  \centerline{\includegraphics[width=4.1cm]{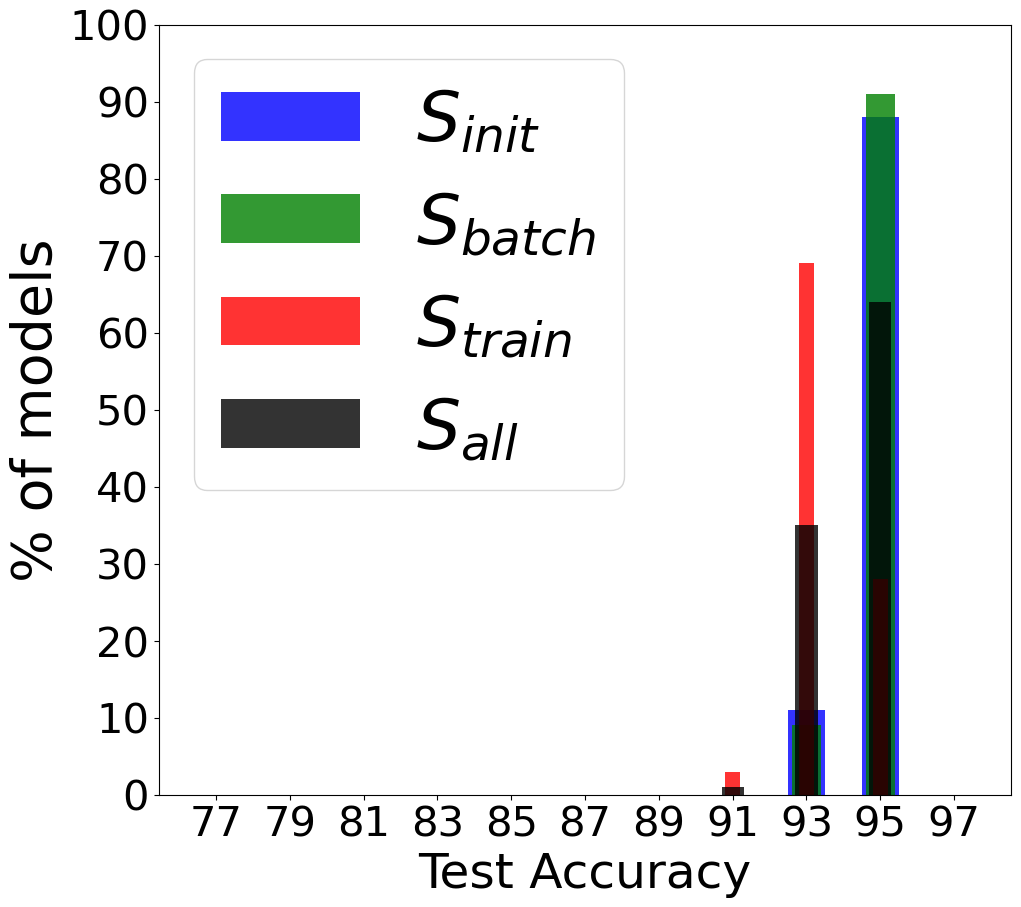}}
\end{minipage}

\caption{ \ninept  Bar plots showing variation in test accuracy and alpha, for Epoch 2 (top row), and Epoch 50 (bottom row).}
\label{fig: scatter plot alpha test accuracy}
\end{figure}

\section{CONCLUSION AND FUTURE WORK}
In this work, we highlighted a different approach to analyzing variability in deep neural networks. Our proposed framework is based on a robust two-sample hypothesis testing problem that uses impartial trimming of the empirical CDF of samples obtained from the logit gap function. The purpose of this test is to assess the similarity/dissimilarity between candidate models in a pool with their consensus (leave-one-out ensemble) model by down-weighting that part of the data that has a greater influence on the dissimilarity. We also provide some evidence that our new measure, the trimming level $\hat{\alpha}$, could be a more informative metric to assess model performance over the commonly used test accuracy.

While in this paper we describe the methodology and an example on a small model, future extensions of this work include applications to very large deep net models and extensions to multi-class classification. Another direction is to explore two-sample hypothesis testing based on other distance metrics like the Wasserstein metric. In this work, we chose to focus on samples from the logit gap function as our probe to understand deep net variability. We can look at other functions of the trained models such as eigen-distribution of the Jacobian or the Neural Tangent Kernel of functions learned by these models. The eigen-distribution of these matrices can provide us information on how well individual candidate models can generalize to test data.

\label{sec: Conclusion and Future Work}

\clearpage

\ninept
\bibliographystyle{IEEEbib}
\bibliography{strings,refs}

\clearpage

\end{document}

%% file: papermacros.tex
\newcommand{\cX}{{\mathcal X}}
\newcommand{\cY}{{\mathcal Y}}

\def\mc{\mathcal}

\def\msf{\mathsf}

\DeclarePairedDelimiter\parens{\lparen}{\rparen}  
\DeclarePairedDelimiter\abs{\lvert}{\rvert}
\DeclarePairedDelimiter\norm{\lVert}{\rVert}

\DeclarePairedDelimiter\braces{\lbrace}{\rbrace}

\newcommand{\R}{\mathbb{R}}

\newcommand{\Indic}[1]{\mathbbm{1}\parens{#1}} 

\DeclareFontFamily{U}{matha}{\hyphenchar\font45}
\DeclareFontShape{U}{matha}{m}{n}{
      <5> <6> <7> <8> <9> <10> gen * matha
      <10.95> matha10 <12> <14.4> <17.28> <20.74> <24.88> matha12
      }{}
\DeclareSymbolFont{matha}{U}{matha}{m}{n}
\DeclareFontSubstitution{U}{matha}{m}{n}
\DeclareFontFamily{U}{mathx}{\hyphenchar\font45}
\DeclareFontShape{U}{mathx}{m}{n}{
      <5> <6> <7> <8> <9> <10>
      <10.95> <12> <14.4> <17.28> <20.74> <24.88>
      mathx10
      }{}
\DeclareSymbolFont{mathx}{U}{mathx}{m}{n}
\DeclareFontSubstitution{U}{mathx}{m}{n}
\DeclareMathDelimiter{\vvvert}{0}{matha}{"7E}{mathx}{"17}
\DeclarePairedDelimiterX{\matnorm}[1]
  {\vvvert}
  {\vvvert}
  {\ifblank{#1}{\:\cdot\:}{#1}}

\NewDocumentCommand{\expect}{ e{_} s o >{\SplitArgument{1}{|}}m }{%
  \operatorname{\mathbb{E}}
  \IfValueT{#1}{{\!}_{#1}}
  \IfBooleanTF{#2}{
    \expectarg*{\expectvar#4}%
  }{
    \IfNoValueTF{#3}{
      \expectarg{\expectvar#4}%
    }{
      \expectarg[#3]{\expectvar#4}%
    }%
  }%
}
\NewDocumentCommand{\expectvar}{mm}{%
  #1\IfValueT{#2}{\nonscript\;\delimsize\vert\nonscript\;#2}%
}
\DeclarePairedDelimiterX{\expectarg}[1]{[}{]}{#1}

\NewDocumentCommand{\prob}{ e{_} s o >{\SplitArgument{1}{|}}m }{%
  \operatorname{\mathbb{P}}
  \IfValueT{#1}{{\!}_{#1}}
  \IfBooleanTF{#2}{
    \probarg*{\probvar#4}%
  }{
    \IfNoValueTF{#3}{
      \probarg{\probvar#4}%
    }{
      \probarg[#3]{\probvar#4}%
    }%
  }%
}
\NewDocumentCommand{\probvar}{mm}{%
  #1\IfValueT{#2}{\nonscript\;\delimsize\vert\nonscript\;#2}%
}
\DeclarePairedDelimiterX{\probarg}[1]{(}{)}{#1}
\renewcommand{\P}{\ensuremath{\mathbb{P}}}
\newcommand{\condon}{\,\ifnum\currentgrouptype=16 \middle\fi|\,} 

\newcommand{\iid}{\overset{\mathrm{i.i.d.}}{\sim}}

